%% file: main.tex
\documentclass[letterpaper, 10pt, conference]{IEEEtran}
\usepackage{times}
\usepackage{xr}
\usepackage{amsmath,amssymb,mathrsfs,gensymb}
\usepackage[]{algorithm2e}
\usepackage{graphicx}
\usepackage{xcolor}
\usepackage{wrapfig}
\usepackage{textcomp}
\usepackage{duckuments}
\usepackage{amsmath,bm}
\usepackage{booktabs}
\usepackage{multirow}

\renewcommand{\arraystretch}{1.2}

\usepackage[sort,numbers,compress]{natbib}
\usepackage{multicol}
\usepackage[bookmarks=true, hidelinks]{hyperref}

\begin{document}

\title{\vspace{5pt}
Robots that redesign themselves through \\ kinematic self-destruction 
}



\author{
\IEEEauthorblockN{%
Chen Yu and
Sam Kriegman
}
\IEEEauthorblockA{%
Northwestern University
}
}

\teaser{
\centering
\includegraphics[trim={0 0 0 0},clip,width=\linewidth]{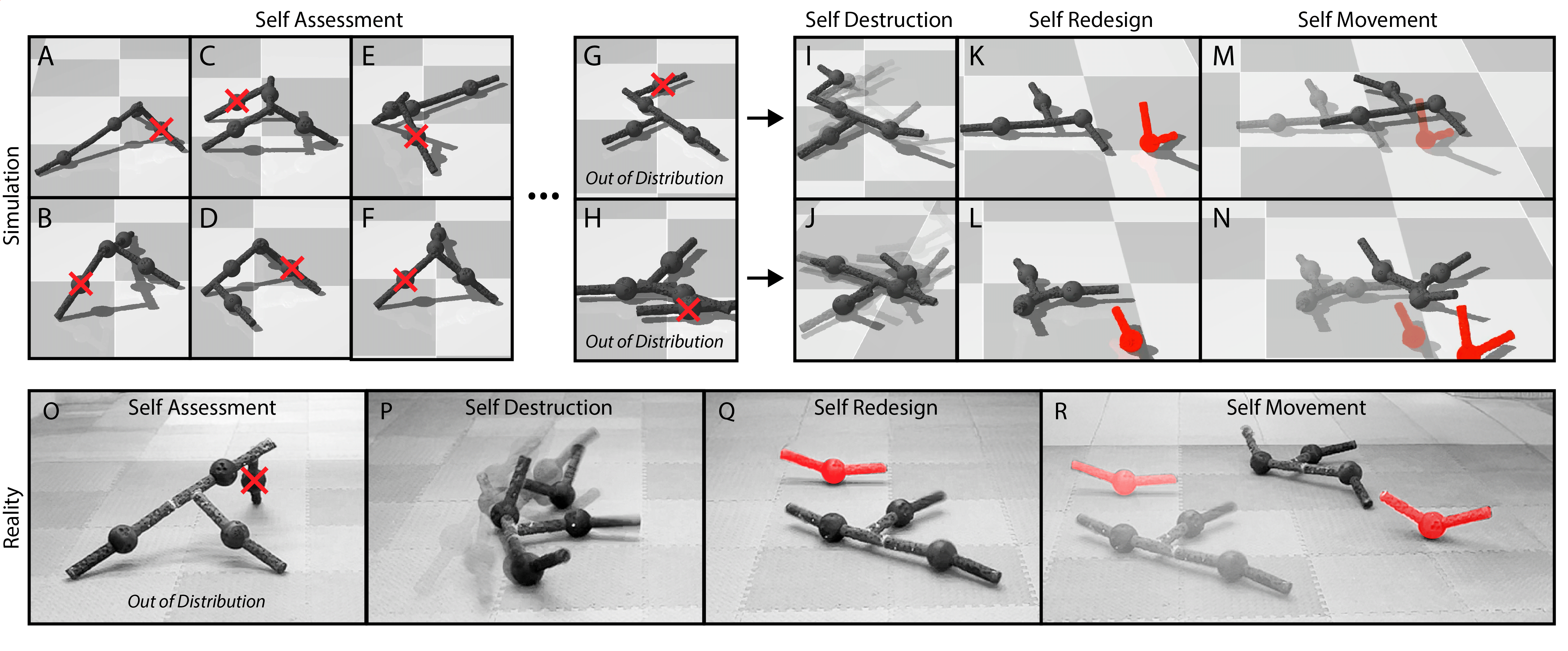} \\
\vspace{-10pt}
    \caption{\textbf{Kinematic self-destruction.}
    Combining the modular building blocks from \cite{yu2025reconfigurable} in different ways yields a diverse set of initial morphologies~\mbox{(\textbf{A-H})}. 
    Unlike \cite{yu2025reconfigurable} however,
    modules were glued rather than bolted together, creating strong yet breakable connections. 
    Using only proprioception feedback, the policy identifies redundant or otherwise undesirable body modules (red Xs in A-H), 
    executes a controlled self-destruction maneuver to mechanically break those unwanted modules off the body (\textbf{I,J}), 
    and then generates forward locomotion in the reduced morphology (\textbf{K-N}).
    After training on a set of simulated morphologies (A-F), the policy generalizes to previously unseen (out of distribution) morphologies in simulation (G,H) and successfully transfers to physical robots with novel initial structures that are likewise out of distribution (\textbf{O-R}).
    Videos and code at \href{https://robot-destruction.github.io/}{\color{blue}\textbf{robot-destruction.github.io}}.
    }
    \label{fig:overview}
    \vspace{-20pt}
}

\maketitle

\input{0_abstract}
\IEEEpeerreviewmaketitle

\input{1_intro}

\input{figures/fig2}
\input{figures/fig3}
\input{figures/fig4}
\input{figures/fig5}

\input{2_methods}
\input{figures/fig6}

\input{3_results}

\input{4_discussion}

\input{5_ack}

\bibliographystyle{unsrtnat}
\bibliography{main}

\end{document}

%% file: 0_abstract.tex
\begin{abstract}


Every robot built to date was predesigned by an external process, prior to deployment. 
Here we show a robot that actively participates in its own design during its lifetime.
Starting from a randomly assembled body, and using only proprioceptive feedback,
the robot dynamically
``sculpts'' itself
into a new design through
kinematic self-destruction:
identifying redundant links within its body that inhibit its locomotion,
and then
thrashing those links against the surface until they break at the joint and fall off the body.
It does so using a single autoregressive sequence model, 
a universal controller that learns in simulation when and how to simplify a robot's body through self-destruction and then adaptively controls the reduced morphology.
The optimized policy successfully transfers 
to reality and generalizes to previously unseen kinematic trees, 
generating forward locomotion that is more effective than otherwise equivalent policies that randomly remove links or cannot remove any.
This suggests that self-designing robots may be more successful than predesigned robots in some cases,
and that
kinematic self-destruction, though reductive and irreversible, could provide a general adaptive strategy for a wide range of robots.

\end{abstract}

%% file: 1_intro.tex
\section{Introduction}
\label{sec:intro}

In nature, several organisms can redesign their bodies through partial self-destruction: intentionally shedding legs, arms, tails, claws and tentacles to evade predators, escape entrapment, or increase their reproductive success~\cite{emberts2017cut}.

If robots could redesign themselves, even through irreversible changes, they would have a much greater capacity to adapt to unanticipated changes in their environment, body, and task.
This additional dimension of adaptation could greatly expand a robot's capabilities and lifespan, which would greatly increase its social utility.
In this paper we explore a special case of a self-designing robot: 
one that cannot grow or absorb new body parts, but that can remove existing parts kinematically through controlled self-destruction.

Most robots are designed from the ground up by human engineers.
Some have been automatically designed~\cite{lipson2000automatic,hornby2003generative,hiller2012automatic,brodbeck2015morphological,cellucci20171d,kriegman2020xenobots,kriegman2020scalable,moreno2021emerge,kriegman2021fractals,kriegman2021kinematic,schaff2022soft,matthews2023efficient,strgar2024evolution,kobayashi2024computational,yu2025reconfigurable},
but the manufactured robot was still given a structure that was predetermined by an external process
and
could not be redesigned in situ without human intervention.
The robot could not, in other words, intentionally modify its own body plan.


Several robots reported to date could change the size of their body parts, but not their overall design.
For example \citet{nygaard2021real} trained a quadruped to extend or partially collapse the length of its four legs to adapt to different terrain types;
the robot could stand up higher or crouch down lower relative to the surface, but it was always a quadruped with the same kinematic tree and surface contacts.
Similarly, \citet{yu2023multi} optimized the length of the legs and torso of a quadruped robot, without changing its overall structure; \citet{kriegman2019automated} optimized the resting volume of each voxel within a soft robot's body.
Other robots have automatically changed their topological structure, but they did so stochastically \cite{white2004stochastic, white2005three, bishop2005programmable, klavins2007programmable} or according to preprogrammed steps~\cite{gilpin2010robot, yang2024self, kurokawa2008distributed, romanishin20153d, davey2012emulating, swissler2018fireant, liang2020freebot}.

Here, in contrast, the robot begins with a randomly assembled kinematic tree, 
evaluates its own structure, 
identifies redundant modules that would hinder locomotion, 
executes a targeted self-destruction
maneuver to remove them,
and initiates forward locomotion
(Fig.~\ref{fig:overview}).

We cast this problem as a sequence modeling task (Fig.~\ref{fig:model}). 
First, we generated a small set of
simulated legged morphologies (Fig.~\ref{fig:model}A-H) composed of modules that break apart if and when the torque at a joint exceeds a threshold. 
Each simulated robot was trained
end-to-end with reinforcement learning, 
learning simultaneously where and how to break off body modules and how to move efficiently in the reduced morphology.
The resulting state-action sequences were distilled into a single transformer model trained across morphologies.
This transformer serves as a universal controller that outputs action sequences, either breaking off modules or moving forward, based solely on proprioceptive observations.

The trained transformer was found to generalize remarkably well across 100 randomly sampled simulated morphologies that were not previously seen during training (e.g.~those in Fig.~\ref{fig:sim-ood}A-I).
The average locomotion speed of these simulated robots was significantly greater than alternative strategies that use the same architecture and training process but cannot self-destruct (Fig.~\ref{fig:sim-ood}) or are forced to destruct in a random way (Fig.~\ref{fig:sim-id}) instead of a self-controlled manner.
Finally, we transferred the trained controller to physical robots, which successfully detached unnecessary modules and moved forward, both with in-distribution designs and on previously unseen (out-of-distribution) morphologies. 
This work thus demonstrates, for the first time, a general-purpose controller that not only enables locomotion across physical robots with a wide range of morphologies but also co-adapts the body and control policy in real time by selectively removing redundant body parts.

%% file: figures/fig2.tex
\begin{figure}[t]
    \centering
    \includegraphics[width=0.49\textwidth]{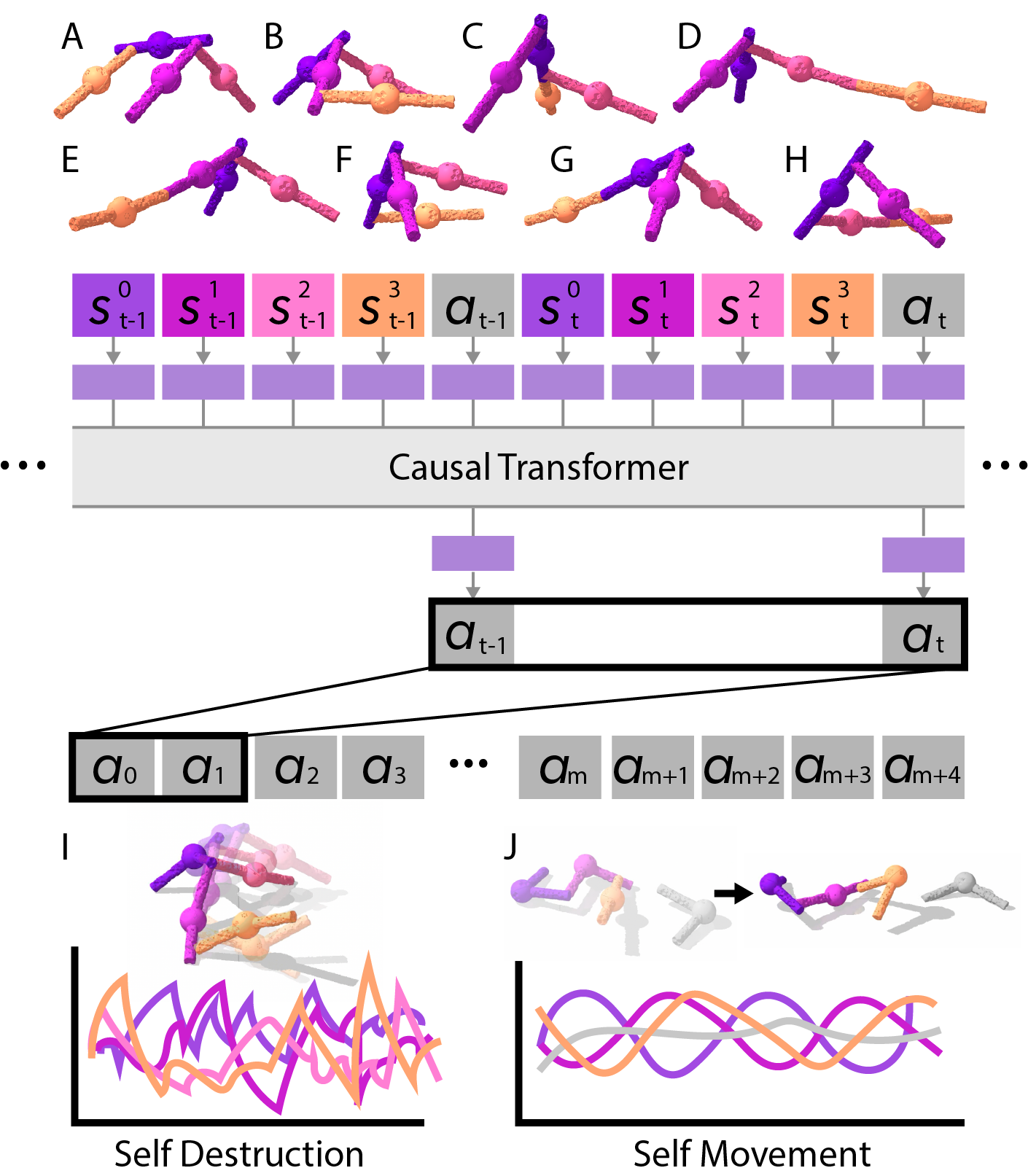}
    \vspace{-15pt}
    \caption{\textbf{Training dataset and transformer architecture.}
    A causal transformer was trained on sensorimotor trajectories collected from eight pre-designed robots (A-H). 
    Each trajectory was flattened into a sequence of per-module states \( \mathbf{s}^i_t \) and actions \( \mathbf{a}_t \). 
    These tokens are processed by the transformer to autoregressively predict the next action, conditioned on the entire state-action history. 
    Inheriting from the expert policies, the output action sequence from the transformer should include a self destruction phase (I) and a self movement phase (J).
    }
    \label{fig:model}
    \vspace{-2em}
\end{figure}

%% file: figures/fig3.tex
\begin{figure*}[th]
    \centering
    \includegraphics[width=1.0\textwidth]{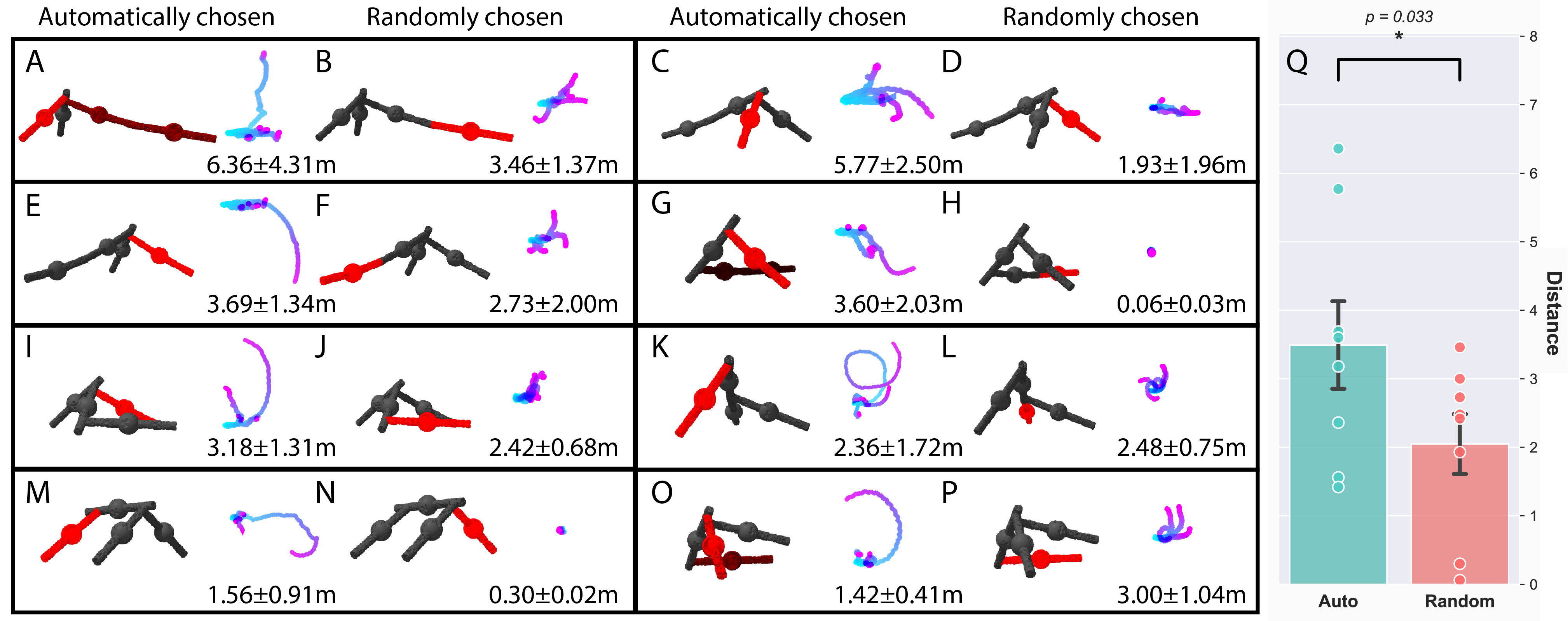}
    
    \vspace{-3pt}
    \caption{\textbf{In-distribution performance in simulation.}
    The policy capable of kinematic self-destruction, in which the expert controller is allowed to autonomously select which module to break, is compared 
    against an otherwise equivalent baseline in which a different module is randomly chosen for removal during expert training and rollout collection. 
    Both sets of rollouts are used to train a transformer policy via identical training pipelines. 
    (\textbf{A-P:}) 
    For each test morphology, top-down trajectories are visualized in time (from cyan to pink) and mean distance traveled (± std) is displayed, across five independent trials. 
    (\textbf{Q:}) 
    Automatically chosen detachments yield better locomotion in terms of mean displacement across these eight in-distribution robots ($p = 0.033$; one-sided paired $t$-test).
    }
    \label{fig:sim-id}
    \vspace{-1em}
\end{figure*}

%% file: figures/fig4.tex
\begin{figure}[t]
    \centering
    \includegraphics[width=0.5\textwidth]{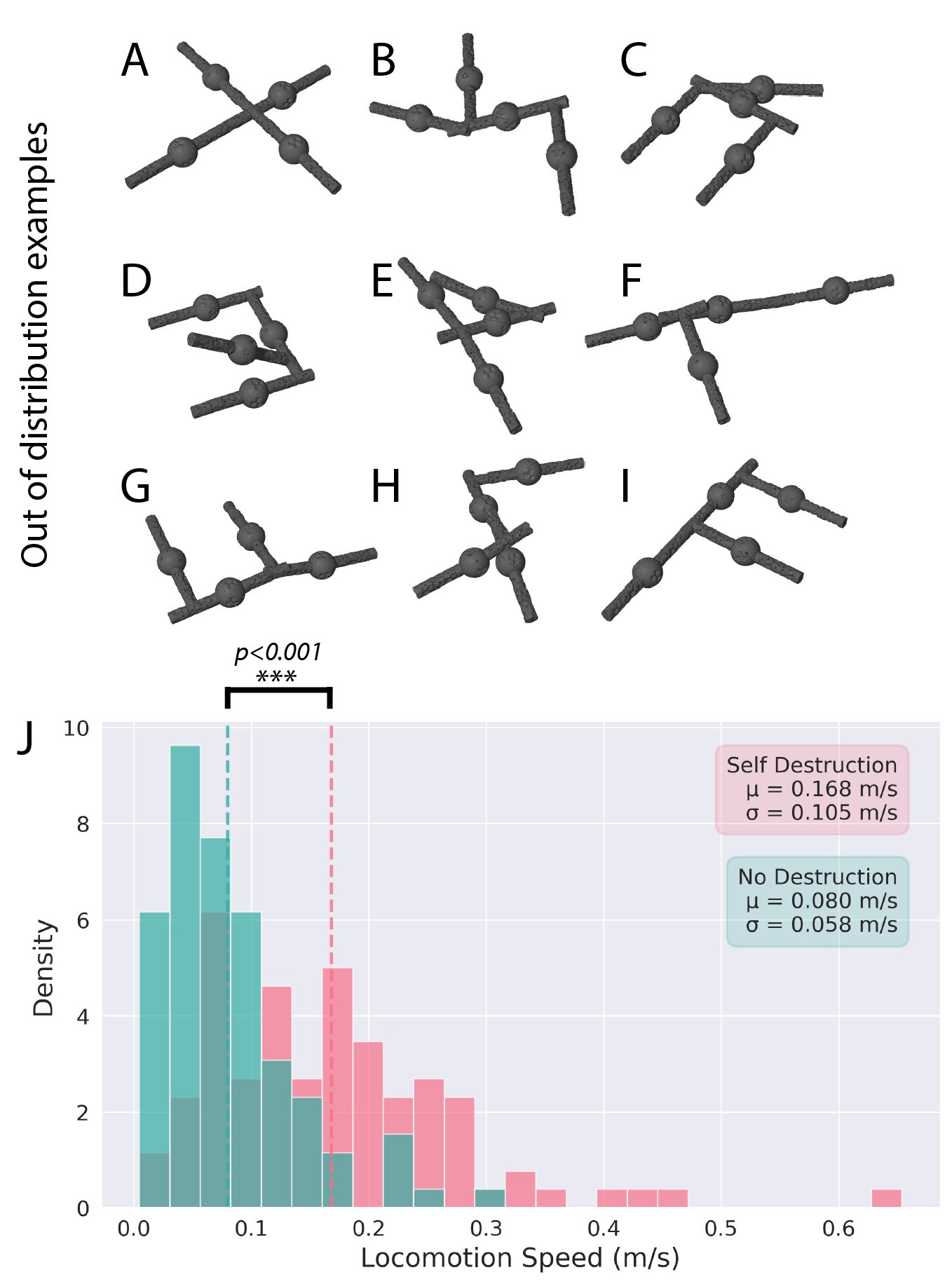}
    \vspace{-18pt}
    \caption{\textbf{Out-of-distribution generalization in simulation.}
  Nine of the 100 previously-unseen morphologies were randomly sampled from the test set (\textbf{A-I}). 
  All 100 test morphologies comprise four modules arranged in a unique configuration, 
  posing diverse control challenges. 
  The mean locomotion speed across all 100 test robots, comparing the policy that uses kinematic self destruction to redesign itself before locomotion (red) against a baseline that does not self-destruct (but is otherwise equivalent; green). 
  For both methods, the reported speed corresponds to the best 10-second segment of the rollout. 
  Self destruction facilitated significantly higher mean locomotion speed, indicating a better generalization to novel morphologies.}
  \label{fig:sim-ood}
  \vspace{-1em}
\end{figure}

%% file: figures/fig5.tex
\begin{figure*}[t]
    \centering
    \includegraphics[width=\textwidth]{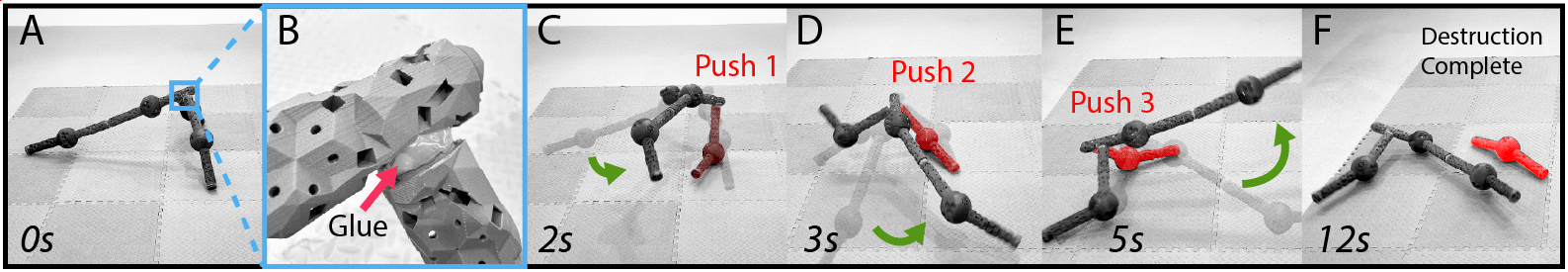}
    \vspace{-16pt}
    \caption{\textbf{A closer look at kinematic self-destruction.} 
    An arbitrary morphology (this one was in the training set) was assembled (\textbf{A}). 
    Modules were glued together to form a breakable bond (\textbf{B}). 
    The policy performs a closed-loop maneuver (with proprioceptive feedback), which in this case consisted of three consecutive pushes (\textbf{C–E}) in which the policy lifted and swung its ``tail'' to exert torque on a specific glued joint. 
    It took 12 seconds to complete this self-destruction (\textbf{F}),
    after which the same policy produces forward locomotion (not pictured) in its redesigned morphology.}
    \vspace{-1em}
    \label{fig:real-id}
\end{figure*}

%% file: 2_methods.tex
\section{Preliminaries}
\label{sec:prelim}

The Transformer model \cite{vaswani2017attention} was developed to process sequential data efficiently. 
At its core, a Transformer consists of multiple stacked self-attention layers interconnected by residual pathways and followed by feedforward networks.

The input to a Transformer is a sequence of tokens, such as words of a language or states and actions of a robot, 
which are first embedded into continuous vector representations. Given an input sequence of token embeddings $(x_1, \dots, x_n)$ of length $n$, each self-attention layer maps it into a new sequence of vectors $\mathbf{z} = (z_1, \dots, z_n)$ of the same length. This is achieved by projecting each input $x_i$ into a query vector $q_i$, a key vector $k_i$, and a value vector $v_i$. The output $z_i$ at position $i$ is then computed as a weighted sum over all value vectors $v_j$, with weights determined by the similarity between $q_i$ and each $k_j$, normalized by a softmax function:

\begin{equation}
z_{i}=\sum_{j=1}^{n} \operatorname{softmax}\left(\frac{\left\{\left\langle q_{i}, k_{j^{\prime}}\right\rangle\right\}_{j^{\prime}=1}^{n}}{\sqrt{d_k}}\right)_{j} \cdot v_{j},
\end{equation}
where $d_k$ denotes the dimensionality of the query and key vectors.

In applications where future information must not influence the current prediction, the attention mechanism is modified to be \textit{causal}. 
This is done by modifying the attention mechanism so that, for each token at position $i$, the model only considers tokens at positions $j \leq i$ when computing the output, ensuring that predictions are based solely on past or current inputs.

\section{Learning self-destruction}
\label{sec:break}

We formulate the ``self-destruct and walk" problem as a Markov Decision Process (MDP), where the action at each timestep specifies the desired position of each joint. The observation space and reward function are detailed below.

\subsection{Observation space}

During expert policy training, the agent receives observations that include the connection status of each module, which is represented as a binary vector of length (N-1) in an N-module system, as well as the projected gravity vector, angular velocity, joint positions and velocities, and the previous action taken by the policy. 

\subsection{Reward function}
\label{sec:rwd}

To encourage purposeful locomotion while allowing for morphological adaptation, we design a reward function that balances net displacement, trajectory efficiency, and retention of active module connections.

Let \( \{\mathbf{x}_t\}_{t=0}^T \subset \mathbb{R}^2 \) denote the 2D position of the robot at each timestep, computed as the average position of the largest physically connected module cluster. 
We maintain a fixed-length position history window of size \( w \), and define the window at timestep \( t \) as \( \mathcal{X}_t = \{\mathbf{x}_{t-w+1}, \ldots, \mathbf{x}_t\} \). 
The net displacement is computed as:
\[
d_t = \|\mathbf{x}_t - \mathbf{x}_{t-w+1}\|_2,
\]
and the path length is the accumulated stepwise distance:
\[
l_t = \sum_{k=t-w+1}^{t-1} \|\mathbf{x}_{k+1} - \mathbf{x}_k\|_2.
\]
We define the average speed as:
\[
s_t = \frac{d_t}{w \cdot \Delta t},
\]
where \( \Delta t \) is the simulation timestep. 
The trajectory efficiency is given by:
\[
e_t = 0.01 \cdot \frac{d_t}{l_t + \varepsilon},
\]
where \( \varepsilon = 10^{-6} \) prevents division by zero.

To discourage excessive destruction, we introduce a regularization term based on active module connections. 
Let \( n_c \) be the expected number of connections in the original morphology, and \( n_a \) be the number of currently active constraints. 
The connection reward is defined as:
\[
c_t = -0.2 \cdot |n_a - n_c|.
\]

The final reward at timestep \( t \) is:
\[
r_t = s_t + e_t + c_t.
\]
This encourages fast, straight-line locomotion while gently penalizing unnecessary module loss, allowing the agent to drop parts only when it improves control performance.


In our experiment, we set a window size \(w=100\), simulation timestep \( \Delta t = 0.05 \) sec, and the expected number of connections \( n_c=2 \).
We train a set of diverse hand-designed configurations, shown in Fig.~\ref{fig:model}A-H, with the MDP described above using SimBa~\cite{lee2025simba}.

\section{Transformer policy}
\label{sec:policy}

Rather than directly deploying the expert policies trained with reinforcement learning on the MDP described in Sect.~\ref{sec:break}, we distill them into a single, general-purpose controller that can operate across diverse morphologies.
In this section, we formulate the problem (Sect.~\ref{sec:mdp2}), describe the model architecture (Sect.~\ref{sec:transformermodel}) and training process (Sect.~\ref{sec:transformertraining}) of this general-purpose controller, and introduce two methods to improve the generalization ability of this controller: Prompt Reset (Sect.~\ref{sec:prompt-reset}) and injecting real world data to the training dataset (Sect.~\ref{sec:realrollouts}).

\subsection{Problem formulation}
\label{sec:mdp2}
We model this universal controller as another MDP.
In this MDP, the state space of the whole robot $\mathcal{S}_\text{robot}$ is the union of per-module sub-state spaces:
\begin{equation}
    \mathcal{S}_\text{robot} = \bigcup_{i=0}^{N} \mathcal{S}_\text{module}^i,
\end{equation}
where \( \mathcal{S}_\text{module}^i \) is the state space of the $i$-th module and \( N \) is the maximum number of modules. 
Each module's state includes projected gravity, angular velocity, cosine of the joint position, and joint velocity. 
The action space \(\mathcal{A}\) consists of the desired joint position for each module, and the reward is the same as described in Sect.~\ref{sec:rwd}.
We then reduce this problem to a sequence modeling task: using a sequential model to fit demonstrated state and action data, produced by the expert policies trained in Sect.~\ref{sec:break}.

\subsection{Model architecture}
\label{sec:transformermodel}

Following \cite{yu2025reconfigurable}, 
we use a causally masked Transformer decoder to autoregressively model state-action trajectories (Fig.~\ref{fig:model}).
At each timestep, the model predicts the next action conditioned on the recent sequence of states and actions.

We represent trajectories as sequences of tokens, where the action \( \bm{a} \in \mathcal{A}\) and each module's sub-state \( \bm{s}^i \in \mathcal{S}_\text{module}^i\) are tokens. Assuming a maximum of \( N \) modules (with \( N = 4 \) in our experiments), a trajectory is encoded as:
\begin{equation}
    \tau = (\bm{s}^0_0, \bm{s}^1_0, \dots, \bm{s}^N_0, \bm{a}_0, \bm{s}^0_1, \dots, \bm{s}^N_1, \bm{a}_1, \dots, \bm{s}^N_T, \bm{a}_T),
\end{equation}
where \( \bm{s}^i_j \) is the state of module \( i \) at timestep \( j \), and \( \bm{a}_j \) is the action at timestep \( j \).
At inference time, we feed the model the most recent \( K \) timesteps (with \( K = 5 \) in our experiments), yielding \( K \times (N+1) \) tokens. Each token is passed through a linear embedding layer with layer normalization. 
Following the Decision Transformer approach \cite{chen2021decision}, we also added timestep-wise positional embeddings, where all tokens at the same timestep share the same positional embedding.

\subsection{Training}
\label{sec:transformertraining}
To train the universal destruction-locomotion policy, we collect batches of expert trajectories generated by the eight hand-designed configurations (pictured in Fig.~\ref{fig:model}A-H; see Sect.~\ref{sec:break}).
We collect \(1 \times10^7\) state-action pairs for each configuration, constructing a pool of sequences of module states and actions over \( K \) timesteps. 
These sequences are used to train the transformer model in an autoregressive fashion.

At each timestep, the transformer is trained to predict the next action token based on the preceding module state and action tokens. 
Specifically, we apply a prediction head to the token corresponding to the final module state at each timestep, and use the output to regress the corresponding action using mean squared error (MSE).

To ensure that the model focuses on physically meaningful predictions, we restrict the loss computation to only those action predictions that belong to the largest physically connected module cluster in each robot at each timestep. 
This allows the model to ignore detached modules, which may not contribute to meaningful control behavior. 
Gradients are thus only propagated through the action tokens corresponding to intact, functional substructures of the robot.

\subsection{Prompt Reset}
\label{sec:prompt-reset}

We observed that when the transformer policy encounters an out-of-distribution state, it may enter a degenerate loop in which it repeatedly outputs nearly identical actions. 
This typically results in the robot becoming frozen, as the model gets trapped in a self-reinforcing cycle of state-action patterns with minimal variation.

To address this, we introduce a simple yet effective \textit{Prompt Reset} mechanism. 
At each timestep \( t \), we monitor the standard deviation of the recent action sequence within a context window of length \( H \), which is set to be the same as the context length used by the transformer model (\( H=K  \) in our experiments). 
Specifically, we compute the maximum standard deviation across all action dimensions over the window \( \{ \mathbf{a}_{t - H + 1}, \dots, \mathbf{a}_t \} \). 
If this maximum standard deviation falls below a threshold \( \tau = 0.2 \), and at least \( t > 50 \) steps have passed, we reset the transformer prompt by clearing the accumulated history of past states and actions. 
This forces the transformer to generate a new sequence from a clean context and helps it escape from frozen or repetitive behavior.


\subsection{Sim-to-real grounding via real-world rollouts}
\label{sec:realrollouts}

To improve out-of-distribution generalization and reduce the sim-to-real gap, we augment our training dataset with real-world trajectories for real-world experiments.
For each simulated robot morphology used during transformer training, we manually construct the corresponding real-world robot in its \textit{final} form, i.e.~after module detachment.
We then apply a generic open-loop controller (a simple phase-shifted sinusoidal oscillator) to execute locomotion in the real world. 
This setup serves as a proxy for innate, reflexive behavior and allows us to capture unstructured but morphology-aware real-world rollouts.
For each morphology, we collect 10 real-world trajectories.

These trajectories are then injected into the transformer training dataset alongside synthetic expert trajectories. 
While the real-world behaviors are open-loop, the transformer absorbs them during training and ultimately converts them into a closed-loop control policy.

To prevent imbalance in the training distribution, we sample real-world data with a fixed probability during training. 
Specifically, at each batch sampling step, there is a 10\% chance of sampling from the real-world data buffer instead of the simulated data. 

%% file: figures/fig6.tex
\begin{figure*}[t]
    \centering
    \includegraphics[width=\textwidth]{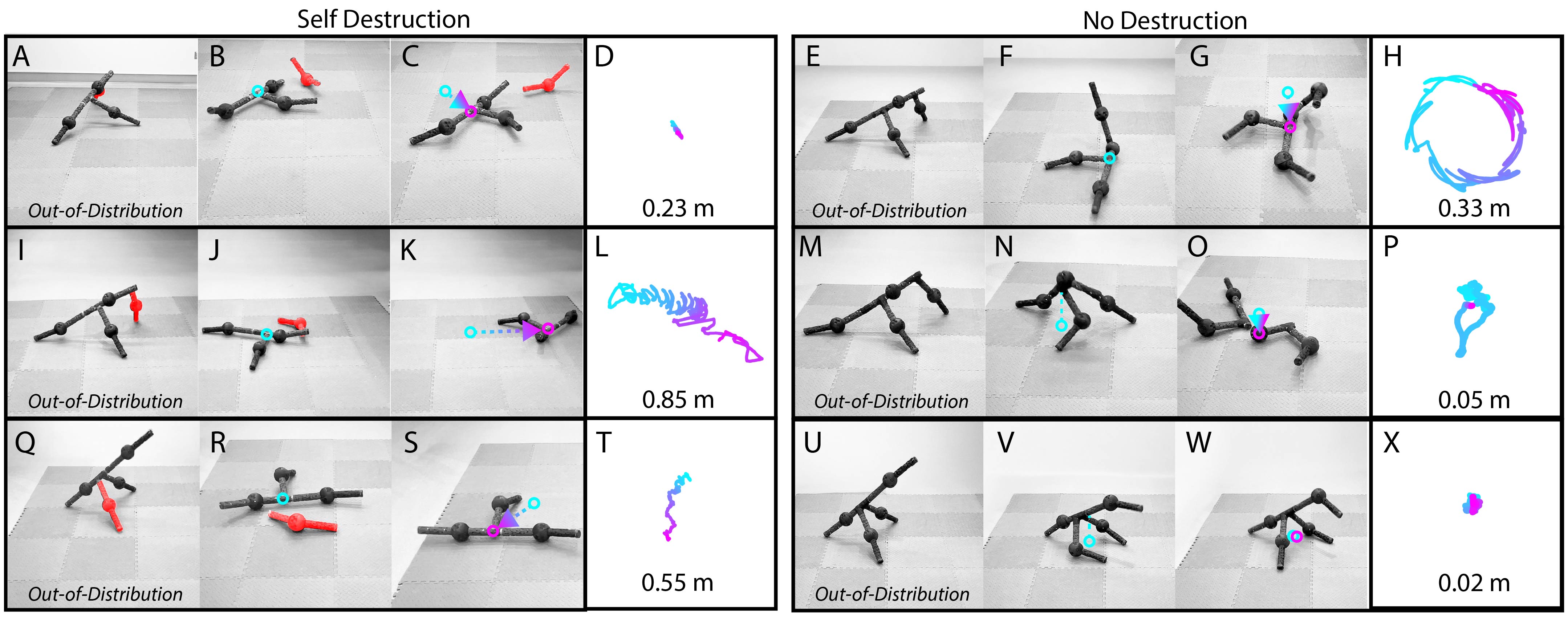}
    \vspace{-18pt}
    \caption{\textbf{Out-of-distribution testing in the real world.}
    Three previously unseen morphologies were assembled and the proposed method was once again compared against the baseline policy, which lacks the ability to self destruct.
    For each case, we show the motion sequence and the torso trajectory for 20 seconds captured via OptiTrack, with color indicating time (from cyan to pink) and net displacement labeled below the trajectory.
    In the first morphology we tested (\textbf{A-H}), although the baseline policy produces a more displacement (F-H), the policy with the ability to self destruct produces a more directional trajectory (B-D); 
    for the last two robots (\textbf{I-P} and \textbf{Q-X}), 
    self destruction consistently redesigns the robot into one with better locomotion performance.}
  \label{fig:real-ood}
  \vspace{-1em}
\end{figure*}

%% file: 3_results.tex
\section{Experimental setup}

Following \cite{yu2025reconfigurable}, we construct modular legged robots. 
But, instead of securing modules with screws, we use hot glue to bond modules at their docking interfaces. 
This allows for rapid assembly and controlled self-destruction.

In simulation, we use MuJoCo's \texttt{equality} constraint system to model rigid connections between modules. 
Specifically, modules are connected via \texttt{weld} constraints, which enforce relative position and orientation across docking joints. 
To allow for physically realistic self-detachment, we monitor the magnitude of the constraint torque in the \( x \) and \( y \) axes (i.e.~bending directions). 
When the combined torque magnitude exceeds a threshold \( \tau_{\text{detach}} \), the corresponding connection is marked as inactive and the constraint is removed:
\[
\text{if } \sqrt{\tau_x^2 + \tau_y^2} > \tau_{\text{detach}} \quad \Rightarrow \quad \text{detach}.
\]
The threshold \( \tau_{\text{detach}} \) is randomized at the beginning of each episode, drawn uniformly from the range \([20, 25]\) Nm to introduce domain variation during training. 
We intentionally ignore torque in the \( z \)-axis (twist) because real-world docking connections exhibit strong resistance to torsional failure. 
Similarly, we omit force thresholds since in practice, torque is the dominant limiting factor for detachment. 
This simplification improves simulation efficiency while capturing the important destruction dynamics.

\section{Results}
\label{sec:results}

To evaluate our approach, we begin by assessing performance on in-distribution morphologies in simulation (Sect.~\ref{sec:sim-id}), where we verify that the policy selects useful modules to remove. We then evaluate generalization by testing on 100 out-of-distribution morphologies in simulation (Sect.~\ref{sec:sim-ood}), and show that our prompt reset mechanism further improves adaptability. Finally, we validate sim-to-real transfer by deploying our transformer controller on two in-distribution physical robots (Sect.~\ref{sec:real-id}) and test on three out-of-distribution real-world morphologies (Sect.~\ref{sec:real-ood}).

\subsection{In-distribution performance in simulation}
\label{sec:sim-id}

Fig.~\ref{fig:sim-id} shows the behavior of our trained transformer policy on a range of in-distribution morphologies in simulation. 
To test the importance of selecting the module the policy automatically chose to discard through self-destruction, 
we compared against a baseline in which, during expert training and rollout generation, a random (but different) module is discarded instead of the one selected by the learned policy.
Both conditions (the original policy and the random baseline) use the same training pipeline. 
Across a wide range of body plans, the modules removed by kinematic self-destruction resulted in more effective locomotion than that of robots with randomly selected modules removed (\(p = 0.033\); one-sided paired t-test).

\subsection{Out-of-distribution performance in simulation}
\label{sec:sim-ood}

To evaluate the generalization capability of our transformer controller, we test it on 100 randomly generated morphologies that were not seen during training  (Fig.~\ref{fig:sim-ood}). 
Each morphology is initialized with a set of fully connected modules, and the policy is tasked with both selecting which modules to detach and producing effective locomotion with the resulting body.
We compare our method against a baseline that uses the same model architecture and training pipeline, but does not allow the robot to drop any modules during expert policy training and throughout execution.
To ensure a fair comparison, we report the average forward speed computed over the best 10-second segment of each trajectory as our method may require several seconds at the start of an episode to evaluate and shed unnecessary modules before initiating locomotion. 
Fig.~\ref{fig:sim-ood}J shows the distribution of trajectory speeds across all 100 morphologies. 

The self-destruction policy achieves a significantly higher mean speed (\( \mu = 0.168~\text{m/s} \), \( \sigma = 0.105 \)) compared to the baseline (\( \mu = 0.080~\text{m/s} \), \( \sigma = 0.058 \)) (\(p<0.001\)). 
The distributions reveal that self-destruction produces a wider range of successful behaviors, including several high-performing outliers beyond \( 0.4~\text{m/s} \), while the baseline struggles to produce any fast locomotion in most cases.

We also conducted an ablation study on a self destruction policy without our prompt reset methods, which has a speed of (\( \mu = 0.131~\text{m/s} \), \( \sigma = 0.088 \)) and is significantly (\( p<0.01\)) worse than the same policy with our prompt reset method.

\subsection{In-distribution testing in the real world}
\label{sec:real-id}

To evaluate how well the transformer controller transfers to physical hardware, we conducted real-world experiments on two in-distribution robot morphologies (Fig.~\ref{fig:model}C and G).
We compared the resulting locomotion performance against the baseline in which the robot cannot self-destruct (described in Sect.~\ref{sec:sim-ood}).
The results are summarized below in Table~\ref{tab:real-world-results}.

Fig.~\ref{fig:real-id} shows snapshots of a learned self destruction strategy from representative trials.
It performs a closed-loop maneuver consisting of three consecutive pushes (Fig.~\ref{fig:real-id}C-E) by lifting and swinging its tail to exert torque on a specific glued joint. 
Despite the presence of real-world imperfections, such as imperfect gluing, the controller repeatedly tried to apply torque on the joint it chose until destruction was complete.


\subsection{Out-of-distribution testing in the real world}
\label{sec:real-ood}

To assess the transferability of the self-destruction policy to the real world, we deploy it on three physical morphologies that were not seen during training and once again compare against the baseline that uses the same architecture and training pipeline but is not allowed to self-destruct (Fig.~\ref{fig:real-ood}). 
Although the baseline generates slightly more displacement for the first morphology we tested, its trajectory is largely circular and inefficient. In contrast, the self destruction policy produces a straighter locomotion path. 
For the other two out-of-distribution physical morphologies we tested, self-destruction outperformed the baseline without destruction both in distance and in producing more directed locomotion trajectories. 

\input{figures/table1}

Table~\ref{tab:real-world-results} summarizes real-world performance across two in-distribution (Fig.~\ref{fig:model}C,G) and three out-of-distribution robot morphologies (Fig.~\ref{fig:real-ood}).
For each morphology, we report the redesign success rate (i.e.~whether the robot successfully detached redundant modules), locomotion success rate (i.e.~whether it sustains locomotion after it starts locomotion), and average forward speed across three trials.
The learned destruction and movement policy achieved a 100\% redesign and locomotion success rate across all tested robots.
While slightly slower in the in-distribution cases, kinematic self destruction consistently resulted in designs that achieved faster locomotion in unseen morphologies.
In contrast, the tested baseline consistently failed to complete the trajectory on out-of-distribution morphologies. Specifically, excessive torque caused structural failure at the bolted sphere-module joint in one out of three trials for the first out-of-distribution robot and two out of three trials for the second.



%% file: figures/table1.tex
\begin{table*}[t]
\centering
\renewcommand{\arraystretch}{2.0} 
\caption{\textbf{Performance comparison across in- and out-of-distribution robots.} Each row shows redesign and locomotion success rates, along with average speed (m/s) across three independent physical trials.}
\vspace{-6pt}
\label{tab:real-world-results}
\begin{tabular}{@{}c c c c c c@{}}
\toprule
\textbf{Robot} & 
\textbf{Distribution} & 
\textbf{Method} &
\textbf{\shortstack{Redesign\\Success Rate}} & 
\textbf{\shortstack{Locomotion\\Success Rate}} & 
\textbf{Avg. Speed (m/s)} \\
\midrule

\multirow{2}{*}{\includegraphics[width=2cm]{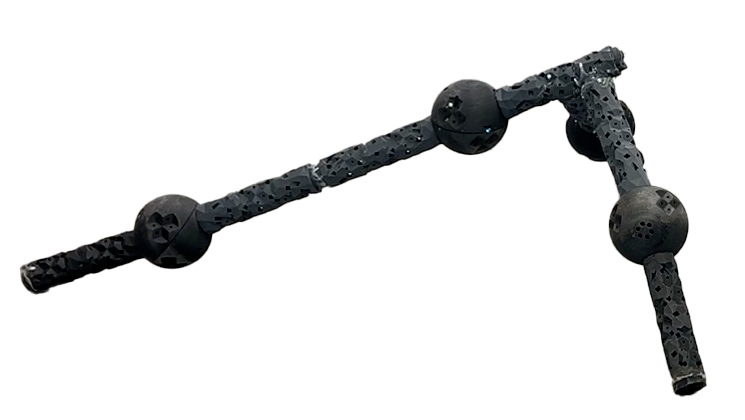}} 
  & \multirow{2}{*}{In-distribution} 
  & Self Destruction     & 100\% & 100\% & 0.051 ± 0.03 \\
  &                      & No Destruction       & N/A & 100\% & \textbf{0.053} ± 0.02 \\
\midrule

\multirow{2}{*}{\includegraphics[width=2.1cm]{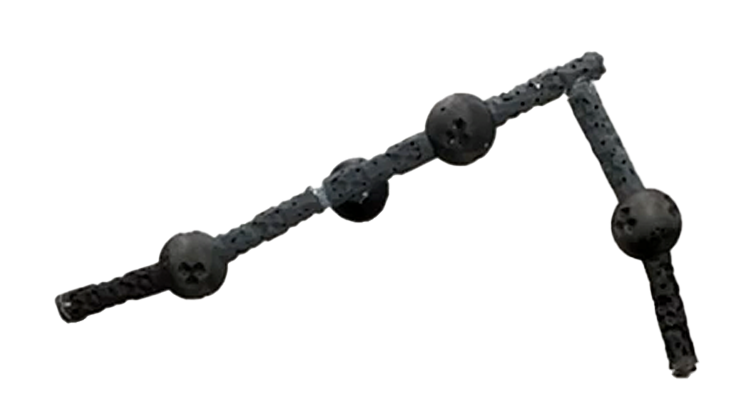}} 
  & \multirow{2}{*}{In-distribution} 
  & Self Destruction     & 100\% & 100\% & 0.022 ± 0.004 \\
  &                      & No Destruction       & N/A & 100\% & \textbf{0.140} ± 0.098 \\
\midrule

\multirow{2}{*}{\includegraphics[width=2.2cm]{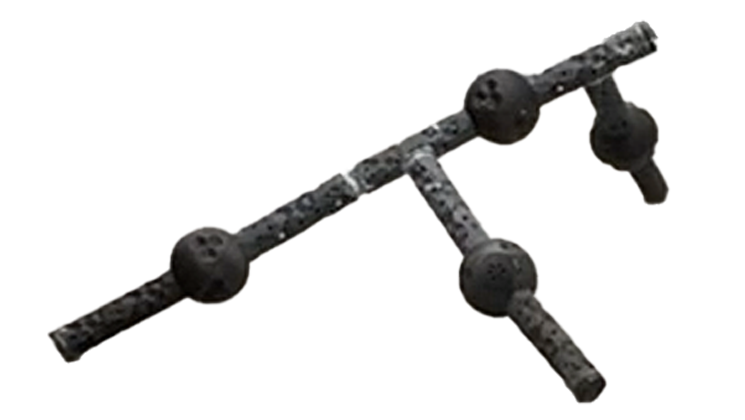}} 
  & \multirow{2}{*}{Out-of-distribution} 
  & Self Destruction     & 100\% & 100\% & \textbf{0.035} ± 0.017 \\
  &                      & No Destruction       & N/A & 66.6\% & 0.027 ± 0.016 \\
\midrule

\multirow{2}{*}{\includegraphics[width=2.4cm]{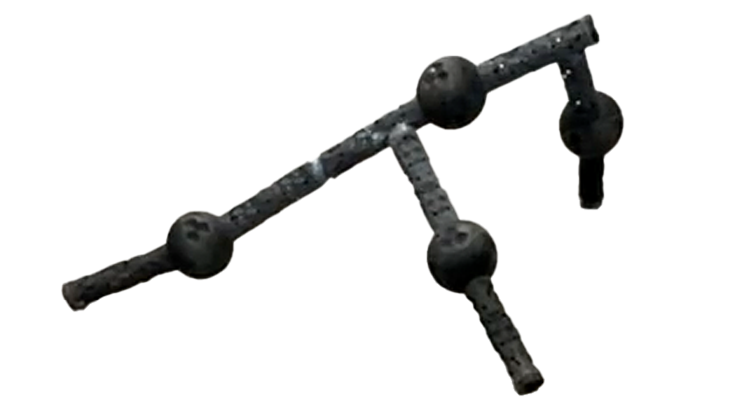}} 
  & \multirow{2}{*}{Out-of-distribution} 
  & Self Destruction     & 100\% & 100\% & \textbf{0.037} ± 0.014 \\
  &                      & No Destruction       & N/A & 33.3\% & 0.032 \\
\midrule

\multirow{2}{*}{\parbox[c]{2.8cm}{\vspace{-1mm}\includegraphics[width=3cm]{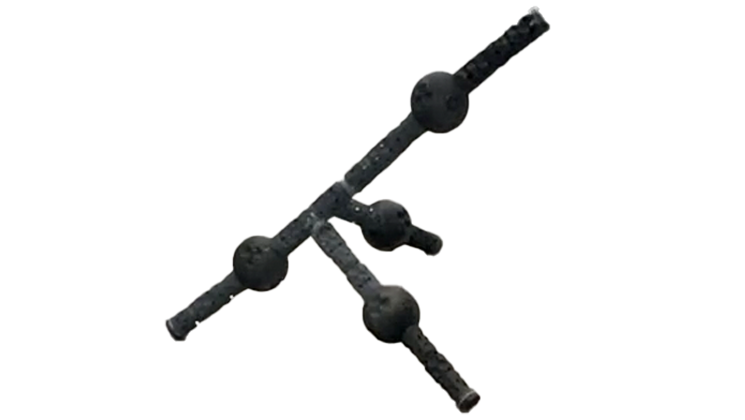}}}
  & \multirow{2}{*}{Out-of-distribution} 
  & Self Destruction     & 100\% & 100\% & \textbf{0.019} ± 0.008 \rule{0pt}{4ex}\\
  &                      & No Destruction       & N/A & 100\% & 0.008 ± 0.004 \rule{0pt}{5ex}\\
\bottomrule
\end{tabular}
\vspace{-1em}
\end{table*}

%% file: 4_discussion.tex
\section{Discussion}
\label{sec:discussion}

In this paper, we introduced the idea of controlled kinematic self-destruction, and showed how robots could use it to redesign themselves to improve their locomotive ability.
We formulated the self-destruction and self-movement problems as a sequence modeling task and distilled expert behaviors with real-world demonstrations into a single autoregressive model.
This closed-loop, transformer-based controller effectively stitched together the 
self-destruction maneuvers (which were distilled from RL policies; Sect.~\ref{sec:break})
and patterns of movement during locomotion
(distilled from open-loop experts; Sect.~\ref{sec:realrollouts}) 
across time.
In initial experiments, however, we observed a peculiar failing of the transformer that trapped certain out-of-distribution robots in a degenerate feedback loop, leading to frozen or repetitive behavior.
We introduced a simple solution to this problem,
\textit{Prompt Reset}, which
clears the accumulated state-action history
if and when diminishing motoric variation is detected, 
thereby avoiding self-referential behavior traps in novel body plans.

The resulting controller generalized to previously unseen robots in both simulation and reality, and outperformed otherwise equivalent baselines that randomly reduced the kinematic tree (random destruction) or could not redesign it (no destruction).
All baselines were architecture-matched (same RL training, same data collection pipeline) 
and differed only in destruction capability.
These results suggest that allowing robots to self-destruct kinematically in a controlled manner could in some cases simplify the control problem and improve the performance and robustness of deployed systems. 


%% file: 5_ack.tex
\section*{Acknowledgments}

This research was supported by
NSF awards 2331581 and 2440412,
and
Schmidt Sciences AI2050 grant G-22-64506.